\documentclass[letterpaper]{article} 
\usepackage{aaai2026}  
\usepackage{times}  
\usepackage{helvet}  
\usepackage{courier}  
\usepackage[hyphens]{url}  
\usepackage{graphicx} 
\usepackage{natbib}  
\usepackage{caption} 
\usepackage{algorithm}
\usepackage{algpseudocode}
\newcounter{linectr} 
\usepackage{amsmath}
\usepackage{amsthm}
\usepackage{bm}
\usepackage{amsfonts}
\usepackage{booktabs}
\usepackage{utfsym}
\newtheorem{definition}{Definition}

\newtheorem{theorem}{Theorem}
\newtheorem{remark}{Remark}
\usepackage{tabularx}
\usepackage{makecell}
\usepackage{multirow}
\usepackage{rotating}
\usepackage{array}

\usepackage{newfloat}
\usepackage{listings}
\DeclareCaptionStyle{ruled}{labelfont=normalfont,labelsep=colon,strut=off} 
\floatstyle{ruled}
\newfloat{listing}{tb}{lst}{}
\floatname{listing}{Listing}
\title{From Pretrain to Pain: Adversarial Vulnerability of Video Foundation Models Without Task Knowledge
}

\author{
  Hui Lu\textsuperscript{\rm 1}, Yi Yu\textsuperscript{\rm 1}\thanks{Corresponding author}, Song Xia\textsuperscript{\rm 1}, Yiming Yang\textsuperscript{\rm 2}, Deepu Rajan\textsuperscript{\rm 2}, Boon Poh Ng\textsuperscript{\rm 2}, Alex Kot\textsuperscript{\rm 1,3}, Xudong Jiang\textsuperscript{\rm 1}
}
\affiliations{
    \textsuperscript{\rm 1}ROSE Lab, School of Electrical and Electronic Engineering, Nanyang Technological University, Singapore\\
    \textsuperscript{\rm 2}Nanyang Technological University, Singapore\\
    \textsuperscript{\rm 3}VinUniversity, Hainoi, Vietnam
    \\
    \{hui007, yuyi0010, xias0002, yiming014\}@e.ntu.edu.sg, \{asdrajan, ebpng, eackot, exdjiang\}@ntu.edu.sg
}

\usepackage{bibentry}

\begin{document}

\maketitle

\begin{abstract}
Large-scale Video Foundation Models (VFMs) has significantly advanced various video-related tasks, either through task-specific models or Multi-modal Large Language Models (MLLMs). 
However, the open accessibility of VFMs also introduces critical security risks, as adversaries can exploit full knowledge of the VFMs to launch potent attacks.
This paper investigates a novel and practical adversarial threat scenario: attacking downstream models or MLLMs fine-tuned from open-source VFMs, without requiring access to the victim task, training data, model query, and architecture.
In contrast to conventional transfer-based attacks that rely on task-aligned surrogate models, we demonstrate that adversarial vulnerabilities can be exploited directly from the VFMs. 
To this end, we propose the Transferable Video Attack (TVA), a temporal-aware adversarial attack method that leverages the temporal representation dynamics of VFMs to craft effective perturbations.
TVA integrates a bidirectional contrastive learning mechanism to maximize the discrepancy between the clean and adversarial features, and introduces a temporal consistency loss that exploits motion cues to enhance the sequential impact of perturbations. 
TVA avoids the need to train expensive surrogate models or access to domain-specific data, thereby offering a more practical and efficient attack strategy.
Extensive experiments across 24 video-related tasks demonstrate the efficacy of TVA against downstream models and MLLMs, revealing a previously underexplored security vulnerability in the deployment of video models.
\end{abstract}

\begin{links}
    \link{Code and appendix}{https://github.com/aloe101/TVA}
\end{links}


\section{Introduction}

Large-scale foundation models trained on diverse datasets have achieved remarkable performance across a variety of tasks such as vision-language chatbots~\cite{achiam2023gpt}, multi-modal reasoning~\cite{li2024llava}, open-world segmentation~\cite{kirillov2023segment}, and spatio-temporal understanding~\cite{li2024mvbench}. Among them, video foundation models~\cite{tong2022videomae,wang2024internvideo2, zhai2023sigmoid}, pretrained on datasets like ImageNet-21k and Kinetics, are well-suited for complex video tasks. Open-source Video Foundation Models (VFMs) offer strong initialization for domain-specific fine-tuning~\cite{liu2024end}, and serve as key components in Multi-modal Large Language Models (MLLMs) and task-oriented frameworks, such as temporal action detection~\cite{zhang2022actionformer} and temporal action segmentation~\cite{chen2024video, wang2020boundary}.

\begin{figure*}
    \centering
    \includegraphics[width=0.95\linewidth]{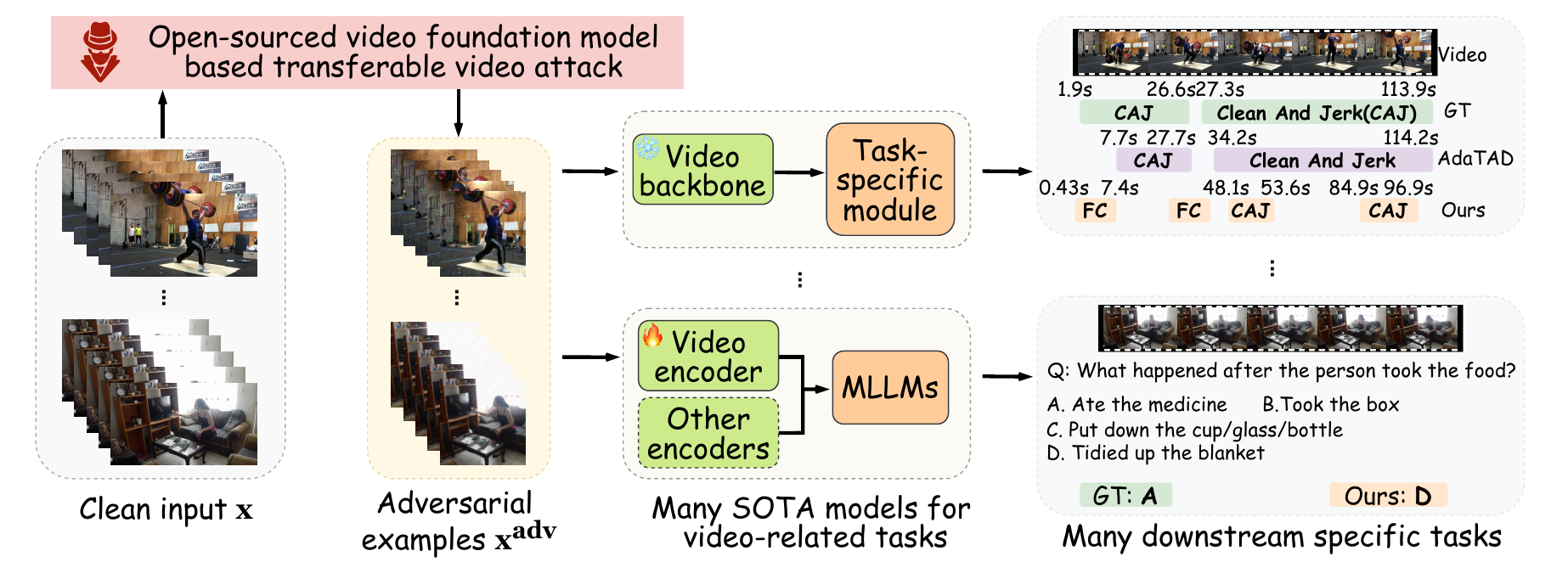}
    \caption{Overview of TVA: TVA deceives various downstream models or MLLMs using only the open-source “Video backbone” or “Video encoder”. “FC” (Frisbee Catch) indicates a misclassification. “AdaTAD” denotes the SOTA model.}
    \label{fig:intro}
\end{figure*}
Despite their strong performance, recent studies~\cite{zhao2023evaluating, schlarmann2023adversarial, xu2023llm,xia2024transferable,lin2024hidemia,lfp1,lfp2,lfp3,lfp4,yu2023backdoor,yu2024purify,yu2024robust,yu2025backdoor,yumtl,wang2024benchmarking,zhou2024securely,zhou2024darksam,zhou2025sam2,zhou2025numbod,wu2025temporal} have raised concerns about the security of deep learning models under adversarial threats. Carefully crafted and nearly imperceptible perturbations (Fig.~\ref{fig:intro}) can mislead well-trained models with high success rates, even when the attacker has limited knowledge, such as access to a surrogate model or restricted query budgets. This vulnerability raises concerns that open-source models, when adapted for downstream use, may inadvertently reveal architectural or parameter-level clues that facilitate adversarial attacks.
Adversarial attacks are commonly divided into white-box~\cite{szegedy2013intriguing,yu2022towards} and black-box~\cite{chen2017zoo} settings, based on whether the attacker can access the victim model’s parameters and architecture. Although white-box attacks assume full transparency, this is often impractical due to deployment and security constraints. Black-box attacks, especially those exploiting transferability, present a more realistic threat, requiring much less knowledge of the target model.

Most transfer-based black-box attacks~\cite{wei2022boosting,wei2023towards,li2025transferable,zeng2025everywhere,yu2025towards} make strong assumptions about the victim model’s task and training distribution. In action recognition, for instance, attackers often assume knowledge of the dataset (\textit{e.g.}, Kinetics-400) and label space, allowing them to train surrogate models with similar properties to generate transferable adversarial examples. However, studies~\cite{zhang2025anyattack,advclip,zhou2023downstream} reveal that such assumptions may not hold in practice due to privacy constraints and data protection laws—particularly when dealing with sensitive biometric data, where models are trained in a self-supervised manner using only features. 
Furthermore, the increasing size and complexity of VFMs exacerbate the difficulty of training such surrogates, limiting the practicality of these methods.

To bridge this gap, we investigate a more practical threat: adversarial attacks without access to the victim model’s task-specific training data, outputs, parameters, full knowledge of the architecture, or the necessity of training a new surrogate model. Considering the prevalent adoption of large VFMs, this study explores the vulnerabilities associated with fine-tuning or directly employing open-source video backbones on downstream datasets. In light of this, we introduce a novel and effective transfer-based method, \textbf{Transferable Video Attack (TVA)}, capable of deceiving downstream task-specific models and MLLMs without prior knowledge of the target task or training datasets. As shown in Fig.~\ref{fig:intro}, our TVA leverages embeddings from open-source VFMs to perform task-agnostic attacks on task-specific models or MLLMs, thereby causing failures across multiple downstream tasks. Specifically, we introduce three complementary components: \textbf{1}) a self-supervised embedding-level attack that induces sparse and consistent perturbations to escape shared decision boundaries; \textbf{2}) a bidirectional temporal-aware contrastive loss that aligns clean and adversarial embeddings in both directions to address perturbation update deviations between surrogate and victim models, correct gradient asymmetry in single-direction contrastive objectives, thereby mitigating surrogate overfitting and enhancing cross-model consistency; and \textbf{3}) a temporal consistency loss designed to disrupt temporal coherence across frames, further amplifying temporal inconsistency.

The contributions of our paper are summarized as follows:
\begin{itemize}
\item We begin an investigation into a more practical while challenging adversarial attack problem: attacking various VFM-based downstream models by solely utilizing the information from the open-sourced foundation model. 
\item We introduce a temporal-aware bidirectional contrastive learning method that enhances attack efficacy by capturing temporal consistency across frames and videos, thereby enabling more effective alignment of temporal representations.
\item We emphasize the critical role of temporal flow dynamics in enhancing adversarial transferability in videos. To this end, we propose a novel temporal consistency loss that exploits sequential continuity to improve performance.
\item Extensive experiments across 24 video-related tasks validate the effectiveness of the proposed approach against VFMs, their downstream models, and MLLMs.
\end{itemize}

\section{Related Work}

\textbf{Adversarial Attacks.} Adversarial attacks are typically categorized into white-box~\cite{szegedy2013intriguing,goodfellow2014explaining} and black-box~\cite{chen2017zoo} settings, based on the attacker's access to the model. White-box attacks leverage gradients of the victim model, while black-box attacks either rely on query feedback~\cite{chen2017zoo} or exploit the transferability of adversarial examples to unseen models.
To improve transferability, optimization-based methods refine gradients to overcome local optima arising from differing decision boundaries across architectures~\cite{liu2016delving}. Representative methods include I-FGSM~\cite{kurakin2018adversarial}, MI-FGSM~\cite{dong2018boosting} with momentum, 
and GRA~\cite{zhu2023boosting}, which averages nearby gradients for smoothing.
Augmentation-based methods diversify inputs to introduce gradient variation. DI~\cite{xie2019improving} uses random resizing and padding; SI~\cite{lin2019nesterov} exploits scale-invariance; TI~\cite{dong2019evading} applies small horizontal and vertical translations to the input;
Admix~\cite{wang2021admix} mixes batch samples, and 
BSR~\cite{wang2024boosting} applies block-wise transformations.
Feature-level attacks further enhance transfer by targeting intermediate representations. FDA~\cite{fda}, FIA~\cite{fia}, NAA~\cite{naa}, and RPA~\cite{rpa} assess neuron importance through activations, gradients, or masking. FTM~\cite{liang2025improving} enhances transferability by mixing attack-specific perturbations with clean features. Besides, X-Transfer~\cite{huang2025x} uses super-ensembles and surrogate scaling, while AnyAttack~\cite{zhang2025anyattack} retrains a surrogate via pretrain+fine-tune. In contrast, we are the first to attack downstream video tasks and MLLMs using open-source VFMs.

\textbf{Temporal Action Detection (TAD).} We adopt TAD as a representative task covering two scenarios: freezing and fine-tuning the foundation model. Our method is broadly applicable to other video-based tasks. TAD aims to identify action categories and their temporal spans in untrimmed videos. As a long-standing challenge in video understanding, TAD has broad applications in areas such as sports analysis and surveillance. Existing methods fall into two categories: feature-based and end-to-end. Feature-based approaches, like ActionFormer~\cite{zhang2022actionformer}, Tridet~\cite{shi2023tridet}, and DyFaDet~\cite{yang2024dyfadet}, operate on pre-extracted features from video foundation models (e.g., I3D~\cite{carreira2017quo}, VideoMAE~\cite{tong2022videomae,wang2023videomae}) using Transformer or CNN architectures. End-to-end methods directly process raw frames, jointly optimizing the encoder and detector~\cite{liu2020progressive}. E2E-TAD~\cite{liu2022empirical} highlighted key design choices and proposed a GPU-efficient baseline. Later improvements include TALLFormer~\cite{cheng2022tallformer} with partial backpropagation and Re2TAL~\cite{zhao2023re2tal} using reversible networks for memory efficiency. AdaTAD~\cite{liu2024end} further scaled to 1B parameters via PEFT.

\textbf{Benchmarks for Video-Involved MLLMs.} The widespread adoption of MLLMs has been accompanied by continuous advancements in video-related benchmarks. For example, SEEDBench~\cite{li2024seed:seedbench} and Video-Bench~\cite{ning2023video:videobench} encompass a diverse range of video-centric tasks designed to thoroughly assess video understanding capabilities. Nevertheless, certain studies have identified a static spatial bias in these benchmarks, stemming from reliance on single frames~\cite{dblei:singlebias}. To address this limitation, MVBench~\cite{li2024mvbench} and Tempcompass~\cite{liu2024tempcompass} introduce video datasets that emphasize temporal features such as motion speed, direction, attribute variations, and event sequencing. Our work uses MVBench and SEEDBench for 23 video-related task comprehensive evaluations.

\begin{figure*}
    \centering
    \includegraphics[width=1\linewidth]{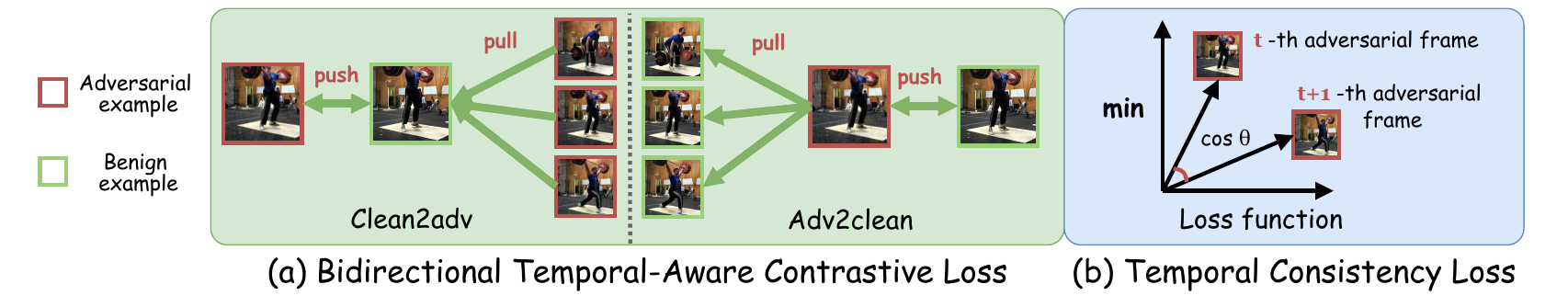}
    \caption{Overview of the Bi-con loss and TC loss: (a) applied to the temporal level, and (b) implemented at the frame level.}
    \label{fig:tva}
\end{figure*}

\section{Preliminary}
\noindent \textbf{Adversarial attacks.}
Let $f$ be a deep model and $\mathcal{L}$ a loss function (e.g., L1 loss). Untargeted adversarial attacks aim to find a perturbation $\boldsymbol{\delta}$ that maximizes the loss under the constraint $\| \bm\delta \|_p \leq \epsilon$, where $\epsilon$ controls imperceptibility. We adopt the $\ell_{\infty}$-norm in this paper. Formally, we have
\begin{equation} \small
\max_{\|\bm\delta\|_{\infty} \leq \epsilon} \; \mathcal{L}(f(\bm{x}), f(\bm{x} + \bm{\delta})).
\label{eq-1}
\end{equation}
A common solution to Eq.~\ref{eq-1} is to iteratively update the perturbation $\boldsymbol{\delta}$ using the loss gradient. A representative method is the Iterative Fast Gradient Sign Method (I-FGSM)~\cite{kurakin2018adversarial}, which updates $\boldsymbol{\delta}$ at each step as:
\begin{equation} \small
\boldsymbol{\delta}_{t+1} = \mathrm{clip}_{\epsilon} \left\{ \boldsymbol{\delta}_t + \alpha \cdot \mathrm{sign} \left( \nabla_{\boldsymbol{\delta}_t} \mathcal{L} (f(\bm{x}), f(\bm{x} + \boldsymbol{\delta}_t)) \right) \right\},
\label{eq-2}
\end{equation}
where $\nabla$ is the gradient operator, $\mathrm{sign}(\cdot)$ computes the element-wise sign, and $\alpha$ is the step size. $\mathrm{clip}_{\epsilon}(\cdot)$ ensures perturbations stay within the $\ell_\infty$-bounded region.

\noindent \textbf{Problem formulation.}
Let $\bm{x} \subset \mathbb{R}^{T \times C \times H \times W}$ be a video sample with $T$ frames of spatial size $H \times W$ and $C$ channels.  
We consider a typical video learning setup, where $f_{\phi_s}$ is a foundation model trained on a large-scale dataset $\mathcal{D}$ and used as a backbone for downstream tasks.  
For each task (e.g., localization, segmentation), a model $f_{\phi_\tau}$ is trained on a dataset $\mathcal{D}_\tau$, initialized from or partially sharing weights with $f_{\phi_s}$.
\begin{equation}\small
\begin{aligned}
    &\phi_s = \arg\min_{\phi_s} \, \mathbb{E}_{(\bm{x}, \bm{y}) \sim \mathcal{D}} \left[ \mathcal{L}(f_{\phi_s}(\bm{x}), \bm{y}) \right], \\
    &\phi_\tau = \arg\min_{\phi_\tau} \, \mathbb{E}_{(\bm{x}_\tau, \bm{y}_\tau) \sim \mathcal{D}_\tau} \left[ \mathcal{L}_\tau(f_{\phi_\tau}(\bm{x}_\tau), \bm{y}_\tau) \right], \\
    &\text{where} ~~\phi_s \overset{\text{share}}{\rightarrow} \phi_\tau ~\text{ or }~ \phi_s \overset{\text{init}}{\rightarrow} \phi_\tau.
\end{aligned}
\end{equation}

\begin{definition}[\textbf{Transferable Adversarial Attack via Open-Sourced Video Foundation Model}]
Given a task-specific video model $f_{\phi_\tau}$ and a clean input $\bm{x}_\tau$, the attacker, without any knowledge of the downstream task or dataset, seeks to craft perturbations $\bm{\delta}_s$ to maximize the loss $\mathcal{L}_{\tau}$, \textit{i.e.}:
\begin{equation} \small
\begin{aligned}
    \!\max_{\|\bm\delta_s\|_{\infty} \leq \epsilon}  \mathcal{L}_\tau \left( f_{\phi_\tau}(\bm{x}_\tau), f_{\phi_\tau}(\bm{x}_\tau \!+\! \bm\delta_s) \right)
    ~\text{s.t.}~   \bm\delta_s \!=\! \mathcal{AT}\!(f_{\phi_s}, \bm{x}_\tau, \mathcal{D}_\tau) ,\nonumber
\end{aligned}
\end{equation}
where $\mathcal{AT}$ denotes the adversarial attack algorithm, $f_{\phi_s}$ is a publicly available video foundation model, and $\mathcal{D}_\tau$ stands for downstream unavailable data. 
\end{definition}
A common approach is to fine-tune a surrogate model $f_{\phi_s^*}$ to mimic the victim model, but this is difficult without access to $\mathcal{D}_\tau$.  
Alternatively, the attacker can directly optimize an attack strategy $\mathcal{AT}^*$ on the source model to maximize the expected output discrepancy of the target model:
\begin{equation} \small
\begin{aligned}
\!\mathcal{AT}^* \!=\! \arg\max_{\mathcal{AT}} \!\mathop{\mathbb{E}}\limits_{\bm{x}_\tau \sim \mathcal{D}_\tau} \!\!\Big[ \mathcal{L}_\tau \!\big( f_{\phi_\tau}(\bm{x}_\tau), f_{\phi_\tau}\big(\bm{x}_\tau \!+\! \mathcal{AT}(f_{\phi_s}, \bm{x}_\tau) \big) \big) \Big].\nonumber
\end{aligned}
\end{equation}
Notably, $f_{\phi_s}$ and $f_{\phi_\tau}$ can be trained on different distributions, $\mathcal{D}$ and $\mathcal{D}_\tau$, with distinct loss functions, leading to a large gap in input-output mappings and gradients. For instance, $\cos(\nabla f_{\phi_s}(\bm{x}_\tau), \nabla f_{\phi_\tau}(\bm{x}_\tau)) \ll 1$, indicating severe gradient misalignment that limits the effectiveness of gradient-based adversarial attacks.

\noindent \textbf{Further analysis on attacking video backbones.}
VFMs are typically adapted to downstream tasks $\tau$ via: (1) fine-tuning with additional modules (e.g., adapters), or (2) freezing the backbone $f_{\phi_{\mathrm{b}}}$ for feature extraction, with task-specific heads built on top of embeddings $\bm{z} = f_{\phi_{\mathrm{b}}}(\bm{x})$. Given the central role of these embeddings and the variability from fine-tuning, a natural attack strategy is to exploit the shared, frozen backbone to craft transferable perturbations:
\begin{equation}\small
    \begin{aligned}
        \!\max_{\|\bm\delta_s\|_{\infty} \leq \epsilon} \mathcal{L}\!\left(f_{\phi^{\tau}_{\mathrm{b}}}(\bm{x}_\tau), f_{\phi^{\tau}_{\mathrm{b}}}(\bm{x}_\tau \!+\! \bm\delta_s)\right) 
        ~\text{s.t.}~\bm{\delta}_s \!=\! \mathcal{A} \mathcal{T}^{*}(f_{\phi_{\mathrm{b}}}, \bm{x}_\tau),\nonumber
    \end{aligned}
\end{equation}
where $\phi^{\tau}_{\mathrm{b}}$ denotes task-specific backbone parameters. Unless stated otherwise, we write $\phi_{\mathrm{b}}$ as $\phi$ and refer to the foundation encoder $f_\phi$ as the surrogate model $f_{\phi_s}$. Our goal is to generate transferable adversarial examples that mislead both fine-tuned backbones and downstream models $f_{\phi_{\tau}}$.

\section{Methodology}
We propose a transferable adversarial attack framework targeting video foundation models and their downstream applications. Our method is composed of three complementary components: \textbf{1)} a self-supervised embedding-level attack to generate base perturbations, \textbf{2)} a bidirectional contrastive objective to improve representation alignment across models, and \textbf{3)} a temporal consistency loss to enhance temporal disruption. We detail each component below.

\subsection{Self-supervised Transferable Adversarial Attack}
We aim to extract the intrinsic vulnerabilities of a frozen video foundation model $f_\phi$ without relying on downstream labels or task-specific outputs. The attack is performed in the embedding space and optimized via the surrogate model.
\begin{definition}[\textbf{Self-supervised Adversarial Perturbation}] \label{d2}
Given a frozen foundation model (feature extractor) $f_{\phi}$ and a video $\bm{x}$ from a downstream dataset $\mathcal{D}_{\tau}$, we extract video embeddings $\bm{z} = f_{\phi}(\bm{x}) \in \mathbb{R}^{T \times D}$, where $D$ is the embedding dimension (e.g., 768 for VideoMAE-base).  
The goal is to generate a perturbation $\bm{\delta}$ such that the adversarial representation $\bm{z}^{adv} = f_{\phi}(\bm{x} + \bm{\delta})$ deviates from $\bm{z}$. Without using task-specific labels or outputs, the attacker optimizes $\bm{\delta}$ via the surrogate $f_{\phi}$ to maximize the embedding-space loss:
\begin{equation} \small 
    \max_{\|\bm\delta_s\|_{\infty} \leq \epsilon} \mathcal{L}(\bm{z},\bm{z}^{adv}).
    \label{eqd2}
\end{equation}
\end{definition}
\noindent Perturbations are updated over $I$ steps using:
$\bm\delta = \text{clip}_{\epsilon}[\bm\delta_0+\sum_{j=1}^I \Delta\bm\delta_j]$,
where $\Delta\bm{\delta}_{j} = \alpha \cdot \text{sign}(\nabla \mathcal{L}(\bm{z}, \bm{z}^{adv}))$ is the update based on $f_{\phi}$. 
Usually, L1 loss is adopted to promote sparse deviations in the embedding space $\mathcal{L}_{\text{L1}}\!=\!\|\bm{z}^{adv}\!-\!\bm{z}\|_1$.

\subsection{Gradient Mismatch between Models}
Although the embedding-level attack enhances transferability, structural differences between models can cause \textbf{gradient mismatch}. To \textbf{explain} this, 
we represent the surrogate $f_{\phi}$ as $m$ sequential blocks $\{f_{\phi_1}^1, \dots, f_{\phi_m}^m\}$ with intermediate features $\{\bm{v}^1, \dots, \bm{v}^m\}$, computed recursively as:
\begin{equation}
\bm{v}^i = f_{\phi_i}^i(\bm{v}^{i-1}), \quad \text{where } \bm{v}^0 = \bm{x}.
\end{equation}
For a downstream task $\tau$, the victim model $f_{\phi_{\tau}}$ may involve parameter updates $\Delta\phi_{\tau}$ to the backbone, and the introduction of new task-specific heads $g_{\psi_{\tau}}$ with updated parameters $\Delta\psi_{\tau}$. Accordingly, the model can take two common forms:
\begin{itemize}
    \item \textbf{Form (a): fined-tuned backbone structure.} The blocks are updated to $\{f_{\phi_1+\Delta\phi_{\tau}^1}^1, \dots, f_{\phi_m+\Delta\phi_{\tau}^m}^m\}$, with each downstream intermediate features $\bm{v}_{\tau}^i$ at block $i$ as:
    \begin{equation}\small
    \!\!\!\bm{v}_{\tau}^0 \!=\! \bm{x}_{\tau},~\bm{v}_{\tau}^i \!=\! f_{\phi_i+\Delta\phi_{\tau}^i}^i\!(\bm{v}_{\tau}^{i-1}) \!=\! f_{\phi_i}^i\!(\bm{v}_{\tau}^{i-1}) \!+\! h_{\Delta\phi_{\tau}^i}^i\!(\bm{v}_{\tau}^{i-1}),
    \end{equation}
    where $h_{\Delta\phi_{\tau}^i}^i$ models the effect of the update $\Delta\phi_{\tau}^i$.
    \item \textbf{Form (b): frozen backbone with task-specific head trained from $\mathcal{D}_{\tau}$.} The backbone $f_{\phi}$ is fixed, and a new head $g_{\psi_{\tau}}$ is appended, yielding the output:
    \begin{equation} \small
        \bm{y}_{\tau} =  g_{\psi_{\tau}}(f_{\phi}(\bm{x}_{\tau})) 
        = g_{\psi_{\tau}}(\bm{z})=g_{\psi}(\bm{z})+h_{\Delta\psi_{\tau}}^{g}(\bm{z}),
    \end{equation}
    where $\bm{z}$ is the extracted embedding. $g_{\psi_{\tau}}\big/g_{\psi}$ models the downstream-specific/original decoder, respectively. $h_{\Delta\psi_{\tau}}^{g}$ captures the influence by the update of encoder.
\end{itemize}

\begin{theorem}[\textbf{Deviation in updating adversarial perturbation}]\label{eqp1}
Let $f_{\phi_{\tau}}$ be the victim model finetuned on task $\tau$. The deviation in perturbation updates between the surrogate and downstream models for \textbf{Form (a)} can be expressed as:
\begin{equation}  \small
    \!\!\!\Delta \bm\delta_{\tau} \!-\! \Delta \bm\delta_{s} \!\leftarrow\! \nabla \mathcal{L} (\bm{y}_{\tau}^{m}) \cdot\! \bigg(\! \prod_{i=1}^{m} \!\left( \nabla f_{\phi^{i}}^{i} \!+\! \nabla h_{\Delta \phi_{\tau}^{i}}^{i} \right) \!-\! \prod_{i=1}^{m} \nabla f_{\phi^{i}}^{i} \!\bigg),
\end{equation}
and, under the \textbf{Form (b)} head-attached case:
\begin{equation} \small
\begin{aligned}
    \Delta \bm\delta_{\tau} - \Delta \bm\delta_{s} \leftarrow \nabla \mathcal{L}(\bm{y}_\tau) \cdot 
\nabla h_{\Delta \psi_\tau}^g(\bm{z}) \cdot \nabla f_\phi(\bm{x}_{\tau}). 
\end{aligned}
\end{equation}
\end{theorem}
\noindent See \textit{Appendix} A.1 for proof of the theorem. Then, we have:
\begin{remark}
Theorem~\ref{eqp1} formally characterizes the deviation in perturbation update dynamics between the surrogate and target models. This deviation arises from the gradient mismatch caused by architectural or parameter discrepancies, particularly the residual transformations $h_{\Delta \phi_\tau^*}^*$ introduced by downstream adaptation. These transformations modify the gradient flow, leading to a misalignment of the optimal adversarial update directions between the surrogate and target models. Consequently, perturbations optimized solely on the surrogate may not align with the decision boundaries of the target model, thereby degrading transferability.
\end{remark}

\subsection{Bidirectional Temporal-Aware Contrastive Attack}
To mitigate this gap, we adopt a contrastive learning strategy to align clean and adversarial embeddings across models. The core idea is to guide adversarial features $\bm{z}^{\text{adv}}$ toward semantically meaningful yet discriminative directions, using clean embeddings $\bm{z}$ as anchors. As shown in Fig.~\ref{fig:tva}(a), this is achieved via a contrastive objective that promotes similarity within positive pairs (clean vs. adversarial of the same frame) and dissimilarity across negatives (different frames). 
Contrastive learning encourages perturbations that shift features away from clean semantics rather than targeting specific task outputs. This results in perturbations that are both task-agnostic and robust to architectural changes, \textbf{thereby mitigating the gradient mismatch} between surrogate and victim models in transferable attacks. 

Given a batch of $n$ samples, let $\bm{z}_{(i)}$ and $\bm{z}_{(i)}^{(\text{adv})}$ denote the clean and adversarial embeddings of the $i$-th sample, respectively. When using clean embeddings as anchors, the standard one-way contrastive loss is defined as:
\begin{equation} \small
\begin{aligned}
\label{eq:con}
\mathcal{L}_{\text{clean} \rightarrow \text{adv}} &= \frac{1}{n} \sum_{i=1}^{n} \mathcal{L}_{\text{clean} \rightarrow \text{adv}}^{(i)},\\
\text{where}~~\mathcal{L}_{\text{clean} \rightarrow \text{adv}}^{(i)}&=
    - \log \frac{\exp(\bm{z}_{(i)} \cdot \bm{z}_{(i)}^{(\text{adv})} / \tau)}{
    \sum_{j=1}^{n} \exp(\bm{z}_{(i)} \cdot \bm{z}_{(j)}^{(\text{adv})} / \tau)}.
\end{aligned}
\end{equation}

\paragraph{Bidirectional Temporal-Aware Contrastive Loss.} 
We can further define another one-way loss using the adversarial embeddings as the anchor:
\begin{equation} \small
\begin{aligned}
\mathcal{L}_{\text{adv} \rightarrow \text{clean}} &= \frac{1}{n} \sum_{i=1}^{n} 
    \mathcal{L}_{\text{adv} \rightarrow \text{clean}}^{(i)},\\
    \text{where}~~\mathcal{L}_{\text{adv} \rightarrow \text{clean}}^{(i)}&=- \log \frac{\exp(\bm{z}_{(i)}^{(\text{adv})} \cdot \bm{z}_{(i)} / \tau)}{
    \sum_{j=1}^{n} \exp(\bm{z}_{(i)}^{(\text{adv})} \cdot \bm{z}_{(j)} / \tau)}.
\end{aligned}
\end{equation}
Then, we introduce a bidirectional loss $\mathcal{L}_{\text{Bi-con}}$ that aligns clean and adversarial features in both directions:
\begin{equation} \small
\begin{aligned}
    &\mathcal{L}_{\text{Bi-con}} = \frac{\mathcal{L}_{\text{clean} \rightarrow \text{adv}}+\mathcal{L}_{\text{adv} \rightarrow \text{clean}}}{2}.
\end{aligned}
\end{equation}

\paragraph{Advantages of $\mathcal{L}_{\text{Bi-con}}$.} $\mathcal{L}_{\text{Bi-con}}$ mitigates \textbf{gradient asymmetry} by treating both clean and adversarial features as anchors, ensuring more balanced updates and improved generalization (Fig.~\ref{fig:tva}). While $\mathcal{L}_{\text{clean} \rightarrow \text{adv}}$ and $\mathcal{L}_{\text{adv} \rightarrow \text{clean}}$ appear symmetric in form, they induce inherently different gradient behaviors during adversarial optimization. As analyzed in Theorem 2, this asymmetry arises from differences in the gradient prefactors with respect to the perturbation $\bm{\delta}_{(i)}$, which can lead to unstable or biased training dynamics. By averaging the two directions, $\mathcal{L}_{\text{Bi-con}}$ resolves this imbalance and stabilizes learning. Moreover, unlike video-level approaches, our frame-level design contrasts each clean frame $\bm{z}_{(i)}$ with all adversarial frames $\bm{z}_{(j)}^{\text{adv}}$ in the batch, expanding negative diversity and enhancing temporal saliency. This design improves transferability across models and tasks without relying on downstream-specific priors.

\begin{theorem}[\textbf{Gradient Asymmetry in Single-direction Contrastive Loss}]
Let $\bm{\delta}_{(i)}$ be the adversarial perturbations applied to input $\bm{x}_{(i)}$ to generate the adversarial feature $\bm{z}_{(i)}^{(\text{adv})}$. 
The gradient of the single-direction contrastive loss from clean to adversarial features w.r.t. $\bm{\delta}_{(i)}$ is given by:
\begin{equation} \small
    \!\!\!\nabla_{\!\bm{\delta}_{(i)}} \!\mathcal{L}_{\text{clean} \to \text{adv}} \!=\! \frac{1}{n\tau} \!\left( \exp\!\left( \!-\!\mathcal{L}_{\text{clean} \to \text{adv}}^{(i)} \right)\! \!-\! 1 \right)\! \bm{z}_{(i)} \cdot \frac{d \bm{z}_{(i)}^{(\text{adv})}}{d \bm{\delta}_{(i)}},
\end{equation}
while the reverse-direction gradient becomes:
\begin{equation} \small
\begin{aligned}
&\!\!\!\nabla_{\!\bm{\delta}_{(i)}} \!\mathcal{L}_{\text{adv} \to \text{clean}} 
\!\!=\!\! \frac{1}{n\tau}\! \!\!\left(\! \!\exp(\!-\!\mathcal{L}_{\text{adv} \to \text{clean}}^{(i)}) \bm{z}_{(i)} \!+\! \sum_{j \ne i} q_j \bm{z}_{(j)} \!\!-\!\! \bm{z}_{(i)} \!\!\right)\! \\
&\cdot \frac{d \bm{z}_{(i)}^{(\text{adv})}}{d \bm{\delta}_{(i)}},
\quad\text{where}~~q_j = \frac{\exp(\bm{z}_{(i)}^{(\text{adv})} \cdot \bm{z}_{(j)} / \tau)}{\sum_{k=1}^n \exp(\bm{z}_{(i)}^{(\text{adv})} \cdot \bm{z}_k / \tau)}.
\end{aligned}
\end{equation}
These gradients shares the same Jacobian term $\frac{d \bm{z}_{(i)}^{(\text{adv})}}{d \bm{\delta}_{(i)}}$, but differ in the prefactor. Thus, we have:
\begin{equation} \small
    \nabla_{\bm{\delta}_{(i)}} \mathcal{L}_{\text{clean} \to \text{adv}}
    \ne \nabla_{\bm{\delta}_{(i)}} \mathcal{L}_{\text{adv} \to \text{clean}}.
\end{equation}
\end{theorem}

\begin{remark}
    In Theorem 2, this asymmetry stems from treating clean or adversarial features as static anchors. When the victim model $f_{\phi_\tau}$ diverges from the surrogate $f_\phi$, this one-way setup yields misaligned perturbations. 
\end{remark}
To address this, we use a bidirectional contrastive loss where clean and adversarial features jointly define the embedding space, \textbf{yielding balanced gradients and better transferability}. As shown in Fig.~\ref{fig:tva}(a), at the frame level, each clean frame $\bm{z}_{(i)}$ contrasts with all adversarial frames $\bm{z}_{(j)}^{\text{adv}}$ in the batch, increasing negative diversity and revealing temporally salient frames.

\subsection{Temporal Consistency Attack}

While spatial semantics are central to adversarial attacks, temporal consistency is equally important in video understanding tasks. \citet{kim2023breaking} shows that enforcing temporal continuity significantly improves model stability and performance. For instance, VideoMAE achieves SOTA accuracy in action recognition by applying masked autoencoding over spatiotemporal tokens to preserve coherence.

Inspired by this, we hypothesize that disrupting temporal consistency in the adversarial feature can effectively destabilize downstream models. Breaking smooth frame-to-frame transitions introduces temporal inconsistencies that impair both spatial understanding and temporal reasoning, thereby improving attack transferability. In light of this, we introduce a temporal consistency loss that penalizes similarity between adjacent adversarial frame embeddings. As shown in Fig.~\ref{fig:tva}(b), given a $T$-frame video, we compute the cosine similarity between consecutive adversarial features $\bm{z}_t^{\text{adv}}$ for \textit{t}-th frame and $\bm{z}_{t+1}^{\text{adv}}$ for (\textit{t}+1)-th frame:
\begin{equation} \small
\mathcal{L}_{\text{TC}} = \frac{1}{T-1} \sum_{t=1}^{T-1} \left(1 - \cos\left(\bm{z}^{\text{adv}}_t, \bm{z}^{\text{adv}}_{t+1}\right)\right).
\end{equation}

\begin{table*}[ht]\small
\setlength{\tabcolsep}{3.5pt}
\centering
\begin{tabular}{l||ccccc|ccccc}
\toprule
\textbf{Dataset} $\rightarrow$& \multicolumn{5}{c|}{\textbf{Thumos14}~\cite{thumos14}} & \multicolumn{5}{c}{\textbf{Charades}~\cite{sigurdsson2016hollywood}}  \\ \midrule
\textbf{Attack} $\downarrow$, \textbf{Model} $\rightarrow$& ActionFormer & Tridet & DyfaDet & AdaTAD & Avg. & ActionFormer & Tridet & DyfaDet & AdaTAD & Avg.  \\ \midrule
Without attack &  50.40 &49.85 &49.85 &  53.17       &   50.07      & 30.36  & 29.51  & 30.07 & 36.05 &  31.50    \\
\midrule
I-FGSM~\cite{kurakin2018adversarial}  & 7.59 & 6.94& 8.29& 21.08 & 10.98     &  4.64  &  4.15     &  4.00  &10.10  &  5.72 \\
MI-FGSM~\cite{dong2018boosting}  & 6.73  & 6.77 & 7.02 & 19.18 & 9.93     &  5.10        & 4.53 &   4.47  & 10.06   & 6.04 \\
DI-FGSM~\cite{xie2019improving}        &  7.38 &  6.65 & 8.27 & 20.93 & 10.81  &  7.18        & 6.03 &  6.58  & 12.50   & 8.07 \\ 
TI-FGSM~\cite{dong2019evading} & 10.32 & 9.15 & 11.75 & 30.08 & 15.33 & 7.36 & 6.82 & 5.83 & 13.80 &8.45 \\
SIM~\cite{lin2019nesterov} & 7.83 & 7.12 & 8.11 & 19.72 & 10.70 & 5.4 & 5.01 & 4.89 & 9.84 & 6.29\\
BSR~\cite{wang2024boosting} &46.11 & 46.99& 50.30 & 52.50 &48.98 & 23.75 & 22.23 & 22.28 & 31.56 & 24.96 \\
FTM~\cite{liang2025improving} & 3.55& 3.40& 4.60& 14.17& 6.43 & 4.94 & 4.60 & 4.64 & 8.82 & 5.75\\ \midrule
Our TVA + I-FGSM       &  1.00  & 0.63 &  0.96&  10.68 & 3.32 & 2.99 & 3.06 &\textbf{ 3.57} & 8.18  & 4.45\\
Our TVA + MI-FGSM   & \textbf{0.12} & 0.44 & \textbf{0.29} & 4.07 & 1.23 & \textbf{2.84}  &  \textbf{2.87}   &    3.61      & 6.25&  3.89   \\
Our TVA + FTM       &  0.79  & \textbf{0.40} &  0.45&  \textbf{3.05} & \textbf{1.17} & 3.00 & 3.04 & 3.76 & \textbf{5.71} & \textbf{3.88}\\
\bottomrule
\end{tabular}
\caption{Results of transfer-based adversarial attacks on different models. We report the average mAP (\%) $\downarrow$. 
We use the VideoMAE-base as the surrogate model on the temporal action detection task. The data with the strongest attack is \textbf{bold}.}
\label{tb1}
\end{table*}
\begin{table*}[h]\small
\centering
\setlength{\tabcolsep}{8pt}
\begin{tabular}{l||cccccc|c} 
\toprule
\textbf{Task} $\downarrow$, \textbf{Attack} $\rightarrow$& \textbf{I-FGSM} & \textbf{MI-FGSM} & \textbf{DI-FGSM} & \textbf{TI-FGSM} & \textbf{SIM} & \textbf{BSR} &  \makecell{\textbf{Our TVA + }\textbf{MI-FGSM}} \\ 
\midrule
Action Sequence         & 38.04 & 28.19 & 30.43 & 30.43 & 9.78  & 11.96 & \textbf{47.83} \\
Action Antonym          & 23.53 & 23.00 & 18.63 & 17.65 & 12.75 & 14.71  &\textbf{39.22} \\
Fine-grained Action     & 53.09 & 38.50 & 43.21 & 37.04 & 23.46 & 18.52 & \textbf{79.01} \\
Unexpected Action       & 17.31 & 20.00 & 14.42 & 10.58 & 6.73  & 3.85  & \textbf{40.38} \\
Object Interaction      & 37.11 & 35.50 & 37.11 & 28.87 & 7.22  & 12.37 & \textbf{68.04} \\
Scene Transition        & 19.41 & 12.50 & 13.53 & 8.24  & 3.53  & 3.53  & \textbf{52.94} \\
Moving Attribute        & 35.24 & 36.50 & 30.48 & 20.95 & 23.81 & 25.71 & \textbf{57.14} \\
State Change            & 6.25  & 3.00  & 3.75  & 3.75  & 6.25  & 2.50  & \textbf{20.00} \\
Character Order         & 34.57 & 31.00 & 24.69 & 23.46 & 16.05 & 22.22 & \textbf{54.32} \\
Counterfactual Inference& 25.00 & 26.00 & 23.44 & 29.69 & 21.88 & 21.88 & \textbf{40.62} \\
\midrule
\textbf{Avg.} & 29.52 & 25.79 & 24.55 & 22.74 & 16.76 & 15.76 & \textbf{42.10} \\
\bottomrule
\end{tabular}
\caption{Results of various attacks on video-related tasks. We report attack success rate (\%) $\uparrow$, \textbf{Avg.} denotes the average performance over all 20 tasks. The surrogate model is LauguageBind, and the victim model is VideoLLaVA~\cite{lin2023video}.}
\label{tb:mvbench}
\end{table*}

\begin{table*}[htbp] \small
\centering
\setlength{\tabcolsep}{1mm}
\setlength{\tabcolsep}{4.5pt}

\begin{tabular}{>{\centering\arraybackslash}p{2cm}>{\centering\arraybackslash}p{2cm} c||cccccccc|c}
\toprule
\makecell{\textbf{Surr}\textbf{ogate}} & 
\makecell{\textbf{Vic}\textbf{tim}} & 
\textbf{Task} & \textbf{I-FGSM} & \makecell{\textbf{MI-}\textbf{FGSM}} & \makecell{\textbf{DI-}\textbf{FGSM}} & \makecell{\textbf{TI-}\textbf{FGSM}} & \textbf{SIM} & \textbf{BSR} & \textbf{X-T} & \textbf{AA}& \makecell{\textbf{Our TVA}}   \\
\midrule
\multirow{4}{*}{\makecell[c]{LanguageBind\\\cite{zhu2023languagebind}}} & 
\multirow{4}{*}{\makecell[c]{VideoLLaVA\\\cite{lin2023video}}} 
& AR & 46.88 & 6.25 & 46.88 & 37.50 & 18.75 & 28.75 &-&-& \textbf{71.88} \\
& & AP & 37.25 & 0 & 29.41 & 27.45 & 13.73 & 13.73 &-&-& \textbf{66.67} \\
& & PU & 15.79 & 5.26 & 18.42 & 15.79 & \textbf{23.68} & \textbf{23.68} &-&-& 21.05 \\
& & \textbf{Avg.} & 33.31 & 3.84 & 31.57 & 26.91 & 18.72 & 22.05 & -&-&\textbf{53.87} \\
\midrule
\multirow{4}{*}{\makecell[c]{QFormer\\\cite{li2023blip}}} & 
\multirow{4}{*}{\makecell[c]{VideoChat2\\\cite{li2024mvbench}}}
& AR & 58.93& \textbf{66.07} & 44.64 & 39.29 & 23.21&42.86&-&-&57.14 \\
& & AP &  51.92& 25.00 & 42.31 & 48.08 &21.15 &51.92&-&-&\textbf{53.85 }\\
& & PU & 25.00 &\textbf{50.00} & 25.00 &28.12 & 25.00 &25.00&-&-&34.38 \\
& & \textbf{Avg.} & 45.28 & 47.02 & 37.32 & 38.50 & 23.12 & 39.93 & -&-&\textbf{48.42}\\
\midrule
\multirow{4}{*}{\makecell[c]{SigLIP\\\cite{zhai2023sigmoid}}} & 
\multirow{4}{*}{\makecell[c]{LLaVA-NeXT\\\cite{li2024llava}}}
& AR & 38.03 & 67.61 & 40.85&  28.17 & 21.13 & 47.86&32.39 &22.54&  \textbf{76.06} \\
& & AP & 40.00 & \textbf{70.00} & 47.14& 31.43 & 31.43 & 42.86&38.57&52.86& 68.57 \\
& & PU & 32.56 & \textbf{39.53} & 32.56& 20.93 & 30.23 & 32.56 & 25.58 & 25.58 &\textbf{39.53} \\
& & \textbf{Avg.} &  36.86 & 59.05 & 40.18 & 26.84 & 27.60 & 41.09 & 32.18 & 33.66 &\textbf{61.39} \\
\bottomrule
\end{tabular}
\caption{ASR (\%) $\uparrow$ across Tasks on SEEDBench. \textit{AR}, \textit{AP}, and \textit{PU} denote Action Recognition, Action Prediction, and Procedure Understanding, respectively. X-T denotes X-Transfer~\cite{huang2025x}, AA denotes AnyAttack~\cite{zhang2025anyattack}.}
\label{tb:seedbench}
\end{table*}

\subsection{Joint Objective for Transferable Video Attacks}
To craft highly transferable adversarial perturbations, we unify the three complementary objectives into a single loss function. The overall optimization target is defined as:
\begin{equation} \small
    \mathcal{L}_{\text{total}} = \mathcal{L}_{\text{L1}} + \mathcal{L}_{\text{Bi-con}} + \mathcal{L}_{\text{TC}},
\end{equation}
where each term captures a distinct aspect of perturbation effectiveness.
Specifically, $\mathcal{L}_{\text{L1}}$ promotes sparse, consistent deviations to escape shared decision regions. $\mathcal{L}_{\text{Bi-con}}$ aligns clean and adversarial embeddings bidirectionally, reducing surrogate overfitting and improving cross-model consistency. $\mathcal{L}_{\text{TC}}$ breaks temporal continuity between frames, targeting the temporal priors of video models.
Together, these objectives guide perturbations to disrupt spatial, semantic, and temporal cues, yielding attacks with enhanced transferability across downstream tasks and architectures.

\paragraph{Novelty and Advantages.} We introduce a task-agnostic attack framework that targets VFMs without relying on downstream task knowledge, labels, or model outputs. Instead of building task-specific surrogates, the method directly leverages the representational structure of frozen backbones. To improve the transferability, we design two auxiliary loss functions that capture temporal and representational dynamics. \textit{These losses are lightweight, broadly compatible, and can be applied to enhance a wide range of \textbf{existing attack methods}, offering a flexible way to boost performance in diverse video scenarios.}

\section{Experiments}
\begin{table}[t] \small
\setlength{\tabcolsep}{2pt}
\begin{tabular}{ccc|ccccc}
\toprule
L1 & Bi-con & TC & ActionFormer & Tridet & DyfaDet & AdaTAD & Avg. \\ 
\midrule
 $\usym{2713}$    &        &  &  7.59 & 6.94& 8.29& 21.08 & 10.98              \\
$\usym{2713}$ &    $\usym{2713}$    &    & 0.94&1.01&1.63&12.43& 4.00
   \\
 $\usym{2713}$  &        &  $\usym{2713}$  & 5.96 & 6.61& 6.86& 20.15& 9.90
    \\
  $\usym{2713}$  &  $\usym{2713}$      & $\usym{2713}$   &   1.00  & 0.63 &  0.96&  10.68 & 3.32  \\
\midrule
 $\usym{2713}$    &        &  &  6.73 & 6.77 &7.02&19.18&    9.93         \\
  $\usym{2713}$ &    $\usym{2713}$    &    & 0.20&0.26&0.38&4.84& 1.42\\
 $\usym{2713}$  &        &  $\usym{2713}$  &5.46&6.37&6.53&17.91&9.07
    \\
 $\usym{2713}$  &  $\usym{2713}$      & $\usym{2713}$   &   0.12 & 0.44 & 0.29 & 4.07 & 1.23  \\
\bottomrule   
\end{tabular}
\caption{Ablation study for the performance of our methods combined by I-FGSM (top) and MI-FGSM (bottom).}
\label{tb:ab}
\end{table}

\subsection{Experimental Setup}
\noindent \textbf{Evaluation details.} We evaluate on temporal action detection (TAD), MVBench (20 tasks), and SEEDBench (3 tasks), totaling 24 video-related tasks. For TAD, we use the average standard \textbf{mean Average Precision (mAP)} at various Intersection over Union (IoU) thresholds over [0.3:0.7:0.1] for THUMOS14 and [0.1:0.9:0.1] for Charades. For MVBench and SEEDBench, treated as classification tasks, we report \textbf{attack success rate (ASR)}, \textit{i.e.}, the percentage of correct answers flipped by the attack. Except for MVBench, we randomly select 100 videos from each task. The randomness test can be found in \textit{Appendix} A.3.

\noindent \textbf{Implementation details.} 
We adopt the MI-FGSM as our base attack. To ensure imperceptibility, the $\ell_{\infty}$ perturbation bound is fixed at $\epsilon=\frac{8}{255}$ for all experiments. For TAD, we set the attack update iterations $I=4$ with the step size $\alpha=\frac{2}{255}$ for computational efficiency. For other tasks, we use $I=20$ and $\alpha=\frac{1}{255}$. 
More details are in \textit{Appendix} A.7.

\subsection{Main Results}
\noindent \textbf{Temporal action detection. }
Table~\ref{tb1} reports results on TAD, with clean input performance shown first, followed by results under adversarial attacks. Our TVA method generates adversarial examples with superior transferability, consistently degrading performance across diverse downstream models. Note that ActionFormer, Tridet, and DyFaDet are top feature-based TAD models (frozen backbone), while AdaTAD is a state-of-the-art end-to-end approach using fine-tuned VideoMAE. We use the original VideoMAE-Base as the surrogate model to ensure a balanced evaluation. TVA nearly nullifies performance in feature-based models and reduces AdaTAD’s accuracy from 53.17\% to 3.05\%, far outperforming the previous best attack (14.17\%). It delivers the most effective attacks on both THUMOS14 and Charades.

\noindent \textbf{MVbench.}
Table~\ref{tb:mvbench} presents the performance of various adversarial attack methods across ten representative tasks selected from MVBench. Our method consistently outperforms existing baselines by a large margin across all tasks. The average performance over all 20 tasks (Avg.) further confirms the superiority of our approach, reaching 42.10 compared to 29.52 by I-FGSM and 25.79 by MIFGSM. This demonstrates enhanced transferability and robustness of the proposed method. More results are in \textit{Appendix} A.4.

\noindent \textbf{SEEDBench. }
As shown in Table~\ref{tb:seedbench}, our TVA generally achieves competitive or superior average rates across all model pairs,  outperforming baselines like I-FGSM, with more stable and balanced performance (\textit{i.e.}, lower variance). Besides, using SigLIP as a surrogate model, our method can transfer to commercial models, which achieves ASR 48.8\% on Gemini-2.0-flash and 33.3\% on GPT5-mini with $\epsilon=\frac{16}{255}$ on AR. 

\begin{figure}[t]
    \centering
    \includegraphics[width=1.0\linewidth]{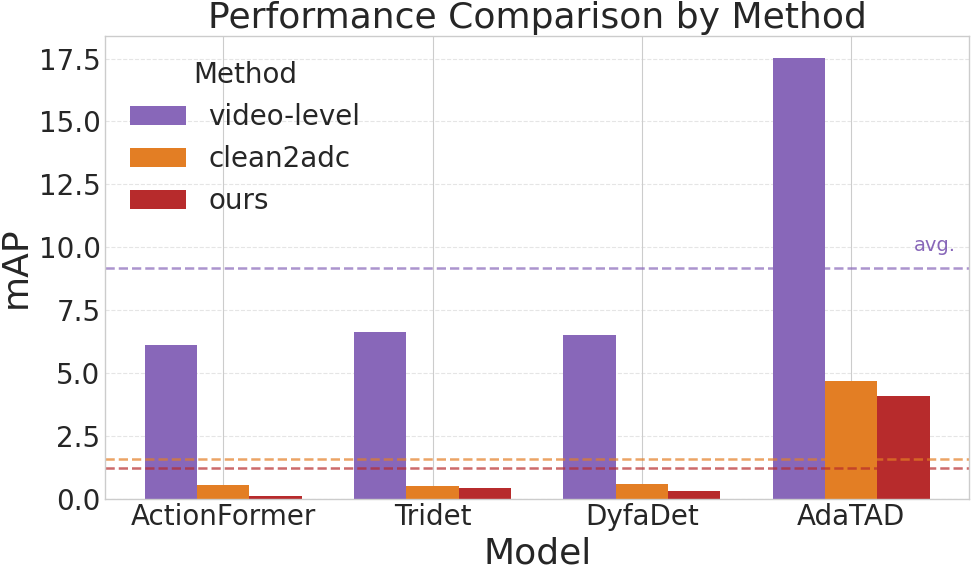}
    \caption{Performance comparison of different contrastive learning strategies on four TAD models. \textit{Video-level} uses standard video-level contrast, \textit{clean2adv}
    applies one-way clean-to-adversarial loss (Eq.~\ref{eq:con}), while \textit{ours} adopts frame-level bidirectional contrast. Our method has the lowest mAP.}
    \label{ab_bicon}
\end{figure}

\subsection{Further analysis}
\noindent \textbf{Ablation.}
Table~\ref{tb:ab} presents an ablation study on three components.  The full model consistently achieves the highest attack performance, indicating the complementary benefits of integrating all three modules. Removing any individual component leads to a drop in performance, with the largest degradation observed when $\mathcal{L}_{\text{Bi-con}}$ is removed. This highlights the critical role of temporal-aware bidirectional contrastive loss in enhancing attack transferability. 

\noindent \textbf{Influence of designs and temperature in Bi-con. }
Fig.~\ref{ab_bicon} shows that our frame-level bidirectional contrastive learning much outperforms both contrastive baselines, demonstrating the importance of fine-grained temporal-aware attack and bidirectional feature supervision. 
Please refer to \textit{Appendix} A.5 for the influence of \textbf{temperature $\bm{\tau}$} in Bi-con.

\noindent\textbf{Resistance to augmentations.} We also investigate our attacks under augmentation-based defenses in \textit{Appendix} A.6.

\section{Conclusion}
Large-scale Video Foundation Models (VFMs) enhance video understanding but introduce new security risks. This paper investigates a more practical and underexplored adversarial threat scenario: transferable adversarial attacks on downstream models fine-tuned or directed from open-source VFMs, without access to the target task, data, or model parameters. We propose \textit{Transferable Video Attack (TVA)}, a task-agnostic framework that exploits the representational structure of VFMs. TVA combines temporal-aware bidirectional contrastive learning with temporal consistency to improve attack transferability. Extensive results across 24 video tasks and many models reveal the vulnerability of VFM-based systems, highlighting the need to reassess their deployment security.

\section{Acknowledgments}
This work was carried out at the Rapid-Rich Object Search (ROSE) Lab, School of Electrical \& Electronic Engineering, Nanyang Technological University (NTU), Singapore. This research is supported by the National Research Foundation, Singapore and Infocomm Media Development Authority under its Trust Tech Funding Initiative. Any opinions, findings and conclusions or recommendations expressed in this material are those of the author(s) and do not reflect the views of National Research Foundation, Singapore and Infocomm Media Development Authority.

\bibliography{aaai2026}

\appendix
\section{Appendix}

\subsection{Proof for Theorem 1}
\begin{proof}
We consider the deviation of perturbation update directions under both the downstream (white-box) and surrogate (transfer) settings. 

In \textbf{Form (a)}, let $f_{\phi_\tau} = (f_{\phi_1}^1, \cdots, f_{\phi_m}^m)$ denote the downstream model after adaptation, and let $h_{\Delta \phi_\tau^i}^i$ denote the residual transformation induced by parameter changes $\Delta \phi_\tau^i$ at the $i$-th layer. Define intermediate features recursively as $\bm{v}_\tau^0 = \bm{x}_\tau$ and $\bm{v}_\tau^i = f_{\phi_i}^i(\bm{v}_\tau^{i-1}) + h_{\Delta \phi_\tau^i}^i(\bm{v}_\tau^{i-1})$.

Then the ideal white-box gradient update of the adversarial perturbation at time $\tau$ is:
\begin{equation} \small
\begin{aligned}
\Delta \bm{\delta}_\tau &\leftarrow \nabla_{\bm{x}_\tau} \mathcal{L}(\bm{v}_\tau^m) \\
&\leftarrow \nabla \mathcal{L}(\bm{v}_\tau^m) \cdot \prod_{i=1}^{m} 
\left( \nabla f_{\phi_i}^i(\bm{v}_\tau^{i-1}) + 
\nabla h_{\Delta \phi_\tau^i}^i(\bm{v}_\tau^{i-1}) \right),
\end{aligned}
\end{equation}
where each Jacobian is computed with respect to the input of that layer. 

In contrast, the surrogate-based update (i.e., without considering downstream adaptation) is given by:
\begin{equation} \small
\Delta \bm{\delta}_s \leftarrow \nabla_{\bm{x}_\tau} \mathcal{L}(f_\phi(\bm{x}_\tau)) 
\leftarrow \nabla \mathcal{L}(\bm{v}_\tau^m) \cdot \prod_{i=1}^{m} \nabla f_{\phi_i}^i(\bm{v}_\tau^{i-1}).
\end{equation}

In \textbf{Form (b)}, we consider a more modular representation. Let the downstream model be decomposed as $g_{\psi + \Delta\psi_\tau}$ and $f_\phi$, where $f_\phi$ is the frozen backbone and $g_{\psi + \Delta\psi_\tau}$ is the task-specific head adapted to the downstream task $\tau$. Let $\bm{z} = f_\phi(\bm{x}_\tau)$ denote the embedding.

Then, the white-box gradient becomes:
\begin{equation} \small
\begin{aligned}
\Delta \bm{\delta}_\tau 
&\leftarrow \nabla_{\bm{x}_\tau} \mathcal{L}(g_{\psi + \Delta \psi_\tau}(f_\phi(\bm{x}_\tau))) \\
&= \nabla_{\bm{z}} \mathcal{L}(\bm{y}_\tau) \cdot \left( \nabla g_{\psi}(\bm{z}) + \nabla h_{\Delta \psi_\tau}^g(\bm{z}) \right) \cdot \nabla_{\bm{x}_\tau} f_\phi(\bm{x}_\tau),
\end{aligned}
\end{equation}
where $\nabla_{\bm{z}} \mathcal{L}$ denotes the derivative with respect to the head input $\bm{z}$.

In contrast, the surrogate-based update (without task adaptation) is:
\begin{equation} \small
\begin{aligned}
\Delta \bm{\delta}_s 
&\leftarrow \nabla_{\bm{x}_\tau} \mathcal{L}(g_{\psi}(f_\phi(\bm{x}_\tau))) \\
&\leftarrow \nabla_{\bm{z}} \mathcal{L}(\bm{y}_\tau) \cdot \nabla g_{\psi}(\bm{z}) \cdot \nabla_{\bm{x}_\tau} f_\phi(\bm{x}_\tau).
\end{aligned}
\end{equation}

Subtracting the two gives:
\begin{equation} \small
\Delta \bm{\delta}_\tau - \Delta \bm{\delta}_s 
= \nabla_{\bm{z}} \mathcal{L}(\bm{y}_\tau) \cdot \nabla h_{\Delta \psi_\tau}^g(\bm{z}) \cdot \nabla_{\bm{x}_\tau} f_\phi(\bm{x}_\tau).
\end{equation}

This expression quantifies the deviation in perturbation updates due to the adapted head $g_{\psi+\Delta\psi_\tau}$.

\vspace{0.5em}
\noindent \textbf{Notation:} 
 $\nabla \mathcal{L}(\bm{v})$ always denotes the gradient with respect to its argument (e.g., $\bm{x}_\tau$ or $\bm{z}$).  
 $h_{\Delta \phi_\tau^i}^i$ and $h_{\Delta \psi_\tau}^g$ respectively model the functional shift due to parameter differences in the $i$-th layer and the head after downstream adaptation. Gradients are evaluated via backpropagation using the chain rule.
\end{proof}

\subsection{Proof for Theorem 2}
\begin{proof}
We derive the gradients of the contrastive losses $\mathcal{L}_{\text{clean} \to \text{adv}}$ and $\mathcal{L}_{\text{adv} \to \text{clean}}$ with respect to the adversarial perturbation $\bm{\delta}_i$.

\paragraph{(1) Gradient of $\mathcal{L}_{\text{clean} \to \text{adv}}$}

Recall the definition:
\[\small
\mathcal{L}_{\text{clean} \to \text{adv}}^{(i)} = - \log \frac{\exp(\bm{z}_i \cdot \bm{z}_i^{(\text{adv})} / \tau)}{\sum_{j=1}^n \exp(\bm{z}_i \cdot \bm{z}_j^{(\text{adv})} / \tau)}. 
\]
Only $\bm{z}_j^{(\text{adv})}$ depends on $\bm{\delta}_i$, and the dependence is non-zero only when $j = i$. Taking the gradient w.r.t. $\bm{\delta}_i$, we get:
\[ \small
    \begin{aligned}
        \nabla_{\bm{\delta}_i} \mathcal{L}_{\text{clean} \to \text{adv}}^{(i)} = 
\frac{1}{\tau} \left( \frac{\exp(\bm{z}_i \cdot \bm{z}_i^{(\text{adv})} / \tau)}{\sum_{j=1}^n \exp(\bm{z}_i \cdot \bm{z}_j^{(\text{adv})} / \tau)} - 1 \right) \bm{z}_i \cdot \frac{d \bm{z}_i^{(\text{adv})}}{d \bm{\delta}_i}.
    \end{aligned}
\]

Using the loss definition again:
\[\small
\frac{\exp(\bm{z}_i \cdot \bm{z}_i^{(\text{adv})} / \tau)}{\sum_j \exp(\bm{z}_i \cdot \bm{z}_j^{(\text{adv})} / \tau)} = \exp(-\mathcal{L}_{\text{clean} \to \text{adv}}^{(i)}),  
\]
we simplify the gradient to:
\[ \small
\nabla_{\bm{\delta}_i} \mathcal{L}_{\text{clean} \to \text{adv}} 
= \frac{1}{n\tau} \left( \exp(-\mathcal{L}_{\text{clean} \to \text{adv}}^{(i)}) - 1 \right) \bm{z}_i \cdot \frac{d \bm{z}_i^{(\text{adv})}}{d \bm{\delta}_i}.
\]
This yields Equation (18).

\paragraph{(2) Gradient of $\mathcal{L}_{\text{adv} \to \text{clean}}$}

Recall the definition:
\[ \small
\mathcal{L}_{\text{adv} \to \text{clean}}^{(i)} = - \log \frac{\exp(\bm{z}_i^{(\text{adv})} \cdot \bm{z}_i / \tau)}{\sum_{j=1}^n \exp(\bm{z}_i^{(\text{adv})} \cdot \bm{z}_j / \tau)}.
\]
This time, the numerator and every term in the denominator depend on $\bm{z}_i^{(\text{adv})}$, and hence on $\bm{\delta}_i$. Taking the gradient:
\[ \small
\nabla_{\bm{\delta}_i} \mathcal{L}_{\text{adv} \to \text{clean}}^{(i)} 
= \frac{1}{\tau} \left( \sum_{j=1}^n q_j \bm{z}_j - \bm{z}_i \right) \cdot \frac{d \bm{z}_i^{(\text{adv})}}{d \bm{\delta}_i},
\]
where
\[ \small
q_j = \frac{\exp(\bm{z}_i^{(\text{adv})} \cdot \bm{z}_j / \tau)}{\sum_{k=1}^n \exp(\bm{z}_i^{(\text{adv})} \cdot \bm{z}_k / \tau)}.
\]
Separating $j=i$ and using the loss definition again:
\[ \small
q_i = \exp(-\mathcal{L}_{\text{adv} \to \text{clean}}^{(i)}),
\]
we rewrite:
\[ \small
    \begin{aligned}
        \nabla_{\bm{\delta}_i} \mathcal{L}_{\text{adv} \to \text{clean}} 
= \frac{1}{n\tau} \left( \exp(-\mathcal{L}_{\text{adv} \to \text{clean}}^{(i)}) \bm{z}_i 
+ \sum_{j \ne i} q_j \bm{z}_j - \bm{z}_i \right) \cdot \frac{d \bm{z}_i^{(\text{adv})}}{d \bm{\delta}_i}.
    \end{aligned}
\]

This gives Equation (19).

\paragraph{(3) Gradient Asymmetry Argument}

Define the direction vectors:
\[ \small
\mathbf{v}_{\text{clean} \to \text{adv}} = \left( \exp(-\mathcal{L}_{\text{clean} \to \text{adv}}^{(i)}) - 1 \right) \bm{z}_i,
\]
\[ \small
\mathbf{v}_{\text{adv} \to \text{clean}} = \exp(-\mathcal{L}_{\text{adv} \to \text{clean}}^{(i)}) \bm{z}_i + \sum_{j \ne i} q_j \bm{z}_j - \bm{z}_i.
\]
In general, unless all $\bm{z}_j$ are colinear with $\bm{z}_i$, the weighted sum $\sum_{j \ne i} q_j \bm{z}_j$ introduces extra components in directions different from $\bm{z}_i$. Therefore,
\[ \small
\mathbf{v}_{\text{clean} \to \text{adv}} \ne \mathbf{v}_{\text{adv} \to \text{clean}},
\quad \Rightarrow \quad
\nabla_{\bm{\delta}_i} \mathcal{L}_{\text{clean} \to \text{adv}} \ne \nabla_{\bm{\delta}_i} \mathcal{L}_{\text{adv} \to \text{clean}}.
\]

This proves the asymmetry of the gradients, which leads to biased adversarial updates if only one direction is considered.
\end{proof}

\begin{table*}[t]
\begin{center}
\begin{minipage}{\textwidth}
  \centering
  \begin{tabular}{lccccc}
\toprule
\textbf{Model} & ActionFormer & Tridet & DyfaDet & AdaTAD & Avg. \\
\midrule
Without attack & $51.15_{(\pm1.26)}$ & $49.54_{(\pm1.21)}$ & $51.00_{(\pm1.25)}$ & $53.45_{(\pm1.44)}$ & $51.01_{(\pm1.38)}$ \\
I-FGSM            & $5.59_{(\pm1.64)}$  & $5.56_{(\pm1.30)}$  & $6.62_{(\pm1.59)}$  & $18.97_{(\pm1.50)}$ & $9.19_{(\pm1.46)}$  \\
MI-FGSM        & $5.41_{(\pm1.32)}$  & $5.60_{(\pm0.84)}$  & $5.80_{(\pm1.12)}$  & $16.78_{(\pm1.78)}$ & $8.40_{(\pm1.23)}$  \\
DI-FGSM       & $7.22_{(\pm0.15)}$ & $7.81_{(\pm0.19)}$ & $8.27_{(\pm0.21)}$ & $20.00_{(\pm0.32)}$ & $10.82_{(\pm0.12)}$ \\
TI-FGSM       & $9.17_{(\pm0.93)}$ & $8.40_{(\pm0.66)}$ & $9.74_{(\pm1.43)}$ & $27.35_{(\pm2.42)}$ & $13.67_{(\pm1.18)}$ \\
SIM           & $6.43_{(\pm1.04)}$ & $6.20_{(\pm0.87)}$ & $7.30_{(\pm0.74)}$ & $16.95_{(\pm2.02)}$ & $9.22_{(\pm1.07)}$  \\
BSR           & $45.87_{(\pm0.61)}$ & $46.69_{(\pm0.71)}$ & $48.62_{(\pm1.45)}$ & $52.53_{(\pm1.29)}$ & $48.43_{(\pm0.58)}$ \\
FTM           & $3.12_{(\pm0.31)}$ & $3.20_{(\pm0.27)}$ & $3.70_{(\pm0.68)}$ & $11.82_{(\pm1.67)}$ & $5.46_{(\pm0.71)}$ \\
\midrule
Our TVA+I-FGSM      & $0.98_{(\pm0.04)}$ & $1.05_{(\pm0.30)}$ & $0.99_{(\pm0.03)}$ & $11.08_{(\pm0.65)}$ & $3.53_{(\pm0.20)}$ \\
Our TVA+MI-FGSM  & $\textbf{0.31}_{(\pm0.19)}$ & $\textbf{0.77}_{(\pm0.34)}$ & $\textbf{0.43}_{(\pm0.12)}$ & $4.50_{(\pm0.55)}$ & $1.50_{(\pm0.30)}$ \\
Our TVA+FTM      & $0.73_{(\pm0.13)}$ & $0.86_{(\pm0.39)}$ & $0.71_{(\pm0.19)}$ & $\textbf{3.62}_{(\pm0.54)}$ & $\textbf{1.48}_{(\pm0.29)}$ \\
\bottomrule
\end{tabular}
\captionof{table}{Experimental randomness of transfer-based adversarial attacks on Thumos14 (subset) on different models. We report mAP(\%) $\downarrow$. We use the VideoMAE-base as the surrogate model on the temporal action detection task. The data with the strongest attack is \textbf{bold}.}
  \label{tb:thumos}
\end{minipage}
\end{center}
\end{table*}
\begin{table*}[t]
\begin{center}
\begin{minipage}{\textwidth}
  \centering
\begin{tabular}{lccccc}
\toprule
\textbf{Model} & ActionFormer & Tridet & DyfaDet & AdaTAD & Avg. \\
\midrule
Without attack & $31.43_{(\pm0.81)}$ & $31.16_{(\pm1.21)}$ & $31.40_{(\pm1.00)}$ & $35.53_{(\pm0.51)}$ & $32.38_{(\pm0.62)}$ \\
I-FGSM            & $5.28_{(\pm0.69)}$  & $4.70_{(\pm0.40)}$  & $5.24_{(\pm1.66)}$  & $10.87_{(\pm1.49)}$ & $6.52_{(\pm0.99)}$  \\
MI-FGSM        & $5.65_{(\pm0.61)}$  & $4.70_{(\pm0.30)}$  & $5.16_{(\pm1.30)}$  & $10.06_{(\pm1.86)}$ & $6.40_{(\pm0.98)}$  \\
DI-FGSM      & $7.35_{(\pm1.40)}$ & $5.90_{(\pm0.67)}$ & $6.50_{(\pm1.08)}$ & $10.94_{(\pm0.88)}$ & $7.67_{(\pm1.00)}$ \\
TI-FGSM      & $7.98_{(\pm0.96)}$ & $7.08_{(\pm0.33)}$ & $6.59_{(\pm0.62)}$ & $14.90_{(\pm1.25)}$ & $9.14_{(\pm0.77)}$ \\
SIM          & $5.94_{(\pm0.57)}$ & $5.28_{(\pm0.36)}$ & $5.36_{(\pm0.75)}$ & $10.10_{(\pm1.46)}$ & $6.68_{(\pm0.76)}$ \\
BSR          & $24.92_{(\pm0.83)}$ & $24.77_{(\pm1.91)}$ & $24.16_{(\pm1.37)}$ & $31.79_{(\pm0.26)}$ & $26.41_{(\pm1.04)}$ \\
FTM          & $4.95_{(\pm0.62)}$ & $4.67_{(\pm0.46)}$ & $4.54_{(\pm0.34)}$ & $9.49_{(\pm1.47)}$  & $5.91_{(\pm0.71)}$ \\
\midrule
Our TVA+I-FGSM     & $3.17_{(\pm0.14)}$ & $3.17_{(\pm0.24)}$ & $3.21_{(\pm0.29)}$ & $7.94_{(\pm0.17)}$  & $4.37_{(\pm0.12)}$ \\
Our TVA+MI-FGSM & $2.87_{(\pm0.03)}$ & $2.93_{(\pm0.13)}$ & $3.08_{(\pm0.39)}$ & $\textbf{6.45}_{(\pm0.60)}$  & $3.83_{(\pm0.09)}$ \\
Our TVA+FTM     & $\textbf{2.85}_{(\pm0.24)}$ & $\textbf{2.71}_{(\pm0.24)}$ & $\textbf{3.05}_{(\pm0.51)}$ & $5.71_{(\pm0.66)}$  & $\textbf{3.58}_{(\pm0.33)}$ \\
\bottomrule
\end{tabular}
\captionof{table}{Experimental randomness of transfer-based adversarial attacks on Charades (subset) on different models. We report mAP(\%) $\downarrow$. We use the VideoMAE-base as the surrogate model on the temporal action detection task. The data with the strongest attack is \textbf{bold}.}
  \label{tb:cha}
\end{minipage}
\end{center}
\end{table*}

\begin{table*}[h]
\centering
\begin{tabular}{l||cccccc|c} 
\toprule
\textbf{Task} & \textbf{I-FGSM} & \textbf{MI-FGSM} & \textbf{DI-FGSM} & \textbf{TI-FGSM} & \textbf{SIM} & \textbf{BSR} &  \makecell{\textbf{Our TVA+}\textbf{MI-FGSM}} \\ 
\midrule
Action Sequence         & 38.04 & 28.19 & 30.43 & 30.43 & 9.78  & 11.96 & \textbf{47.83} \\
Action Prediction       & 37.65 & 38.00 & 32.94 & 30.59 & 12.94 & 16.47 & \textbf{61.18} \\
Action Antonym          & 23.53 & 23.00 & 18.63 & 17.65 & 12.75 & 14.71  &\textbf{39.22} \\
Fine-grained Action     & 53.09 & 38.50 & 43.21 & 37.04 & 23.46 & 18.52 & \textbf{79.01} \\
Unexpected Action       & 17.31 & 20.00 & 14.42 & 10.58 & 6.73  & 3.85  & \textbf{40.38} \\
Object Existence        & 14.15 & 13.13 & \textbf{15.09} & 11.32 & 14.15 & 13.21 & 13.21 \\
Object Interaction      & 37.11 & 35.50 & 37.11 & 28.87 & 7.22  & 12.37 & \textbf{68.04} \\
Object Shuffle          & 37.04 & \textbf{39.50} & 30.86 & 28.40 & 33.33 & 19.75  & 35.80 \\
Moving Direction        & 16.00 & 15.50 & 14.00 & 12.00 & 16.00 & 14.00  & \textbf{18.00} \\
Action Localization     & 25.81 & 15.00 & 20.97 & 24.19 & 20.97 & 14.52  & \textbf{27.42} \\
Scene Transition        & 19.41 & 12.50 & 13.53 & 8.24  & 3.53  & 3.53  & \textbf{52.94} \\
Action Count            & 8.24  & 11.50 & 9.41  & 11.76 & 7.06  & 5.88  & \textbf{18.82} \\
Moving Count            & 43.10 & 32.00 & 32.76 & 18.97 & 39.66 & 24.14 &  \textbf{46.55} \\
Moving Attribute        & 35.24 & 36.50 & 30.48 & 20.95 & 23.81 & 25.71 & \textbf{57.14} \\
State Change            & 6.25  & 3.00  & 3.75  & 3.75  & 6.25  & 2.50  & \textbf{20.00} \\
Fine-grained Pose       & 39.13 & 42.35 & 39.13 & 39.13 & 30.43 & 23.19 & \textbf{57.97} \\
Character Order         & 34.57 & 31.00 & 24.69 & 23.46 & 16.05 & 22.22 & \textbf{54.32} \\
Egocentric Navigation   & 9.23  &\textbf{ 10.50} & 6.15  & 4.62  & 7.69  & 4.62  & 7.69 \\
Episodic Reasoning      & 18.09 & 25.00 & 14.89 & 21.28 & 9.57  & 14.89  & \textbf{34.04} \\
Counterfactual Inference& 25.00 & 26.00 & 23.44 & 29.69 & 21.88 & 21.88 & \textbf{40.62} \\
\midrule
\textbf{Avg.} & 29.52 & 25.79 & 24.55 & 22.74 & 16.76 & 15.76 & \textbf{42.10} \\
\bottomrule
\end{tabular}
\caption{Results of various attacks on video-related tasks. We report attack success rate (\%) $\uparrow$, Avg. denotes the average
performance over all 20 tasks. The surrogate model is LauguageBind, and the victim model is VideoLLaVA. }
\label{tb:mvbench}
\end{table*}

\begin{table*}[h]
\centering
\begin{tabular}{l||cccccc|c}
\toprule
\textbf{Task} & \textbf{I-FGSM} & \textbf{MI-FGSM} & \textbf{DI-FGSM} & \textbf{TI-FGSM} & \textbf{SIM} & \textbf{BSR} & \textbf{Our TVA + MI-FGSM} \\
\midrule
Action Sequence            & 49.66 & \textbf{75.86} & 42.76 & 41.38 & 31.72 & 40.69 & 73.10 \\
Action Prediction          & 50.39 & 72.87 & 45.74 & 38.76 & 24.81 & 46.51 & \textbf{73.19} \\
Action Antonym             & 14.69 & 49.65 & 16.78 & 11.89 & 10.49 & 25.87 & \textbf{52.45} \\
Fine-grained Action        & 37.76 & \textbf{77.55} & 46.94 & 38.78 & 29.59 & 50.00 & 74.49 \\
Unexpected Action          & 23.31 & 36.20 & 23.31 & 17.18 & 15.95 & 23.31 & \textbf{37.42} \\
Object Existence           & 20.33 & \textbf{37.40} & 17.07 & 25.20 & 12.20 & 26.02 & 36.59 \\
Object Interaction         & 44.94 &\textbf{ 71.91} & 46.63 & 34.27 & 26.97 & 45.51 & \textbf{71.91} \\
Object Shuffle             & 10.39 & 14.29 & 10.39 & 7.79  & 3.90  & 10.39 & \textbf{15.58} \\
Moving Direction           & 65.82 & 74.68 & 70.89 & 49.37 & 46.84 & 75.95 & \textbf{77.22} \\
Action Localization        & 50.00 & 68.06 & 47.22 & 38.89 & 43.06 & 43.06 & \textbf{70.83} \\
Scene Transition           & 11.41 & 38.04 & 10.87 & 8.15  & 4.89  & 19.57 & \textbf{38.92} \\
Action Count               & 28.05 & \textbf{34.15} & 29.27 & 30.49 & 28.05 & 28.05 &\textbf{ 34.15} \\
Moving Count               & 48.42 & 58.95 & 42.11 & 37.89 & 36.84 & 50.53 & \textbf{60.55 }\\
Moving Attribute           & 24.66 & 43.84 & 18.49 & 23.29 & 16.44 & 36.30 & \textbf{52.05} \\
State Change               & 27.68 & 34.82 & 26.79 & 23.21 & 19.64 & 24.11 & \textbf{36.61} \\
Fine-grained Pose          & 46.15 & 64.10 & 40.17 & 28.21 & 18.80 & 28.21 & \textbf{65.70} \\
Character Order            & 29.73 & 45.95 & 28.38 & 24.32 & 20.27 & 31.76 & \textbf{47.38} \\
Egocentric Navigation      & 18.31 & 26.76 & 18.31 & 15.49 & 14.08 & 18.31 & \textbf{29.58} \\
Counterfactual Inference   & 38.39 & 53.57 & 41.07 & 39.29 & 32.14 & 52.68 & \textbf{54.46} \\
\midrule
\textbf{Avg.}              & 33.69 & 51.51 & 32.80 & 28.10 & 22.98 & 35.62 & \textbf{52.75} \\
\bottomrule
\end{tabular}
\caption{Results of various attacks on video-related tasks. We report attack success rate (\%) $\uparrow$, Avg. denotes the average
performance over all 20 tasks. The surrogate model is SigLIP, and the victim model is LLaVA-NeXT. }
\label{tb:llavanext}
\end{table*}

\subsection{Randomness test}
We evaluated 10 attack methods presented in our paper over 3 random seed runs on the subset of the representative video downstream task--Temporal Action Detection (TAD), and reported the mean performance with its standard deviation. We use the same experimental setting provided in the Section. \textit{Main results} except for the random seeds. The results indicate that the randomness of our TVA is relative small among all attacking methods. As shown in Table~\ref{tb:thumos} and ~\ref{tb:cha}, our approach consistently achieves superior performance compared to all existing methods, \textit{i.e.} stronger attack and lower standard deviation, demonstrating the robustness and stability of our proposed attack.

\subsection{The performance of TVA on MVBench}
Table~\ref{tb:mvbench} reports the attack success rates (\%) of various methods across all 20 tasks in MVBench. Our method achieves the highest average performance of \textbf{42.10\%}, significantly outperforming the strongest baseline MIFGSM (25.79\%). TVA ranks first on 17 out of 20 tasks, with notable improvements on fine-grained tasks such as \textit{Fine-grained Action} (79.01\%), \textit{Object Interaction} (68.04\%), and \textit{Action Prediction} (61.18\%). These results highlight TVA’s superior transferability and effectiveness in black-box attack settings without relying on downstream task supervision.

Table~\ref{tb:llavanext} presents complete results of TVA across the 20 MVBench tasks when transferring adversarial examples from SigLIP to LLaVA-NeXT. TVA achieves the highest overall average attack success rate of \textbf{52.75\%}, outperforming all compared baselines, including the strongest competitor MIFGSM (51.51\%) as well as transformation-based methods like TI-FGSM (28.10\%), SIM (22.98\%), and BSR (35.62\%). TVA ranks first on the majority of tasks (17/20), showing particularly large gains on tasks involving temporal reasoning, such as \textit{Action Prediction}, \textit{Moving Direction}, and \textit{Action Localization}. Although the improvement is less pronounced on a few tasks like \textit{Object Shuffle}, TVA consistently maintains strong performance across both spatial and reasoning-heavy tasks, demonstrating its superior transferability and robustness in transferable attack settings. 

\begin{figure}[t]
    \centering
    \includegraphics[width=1\linewidth]{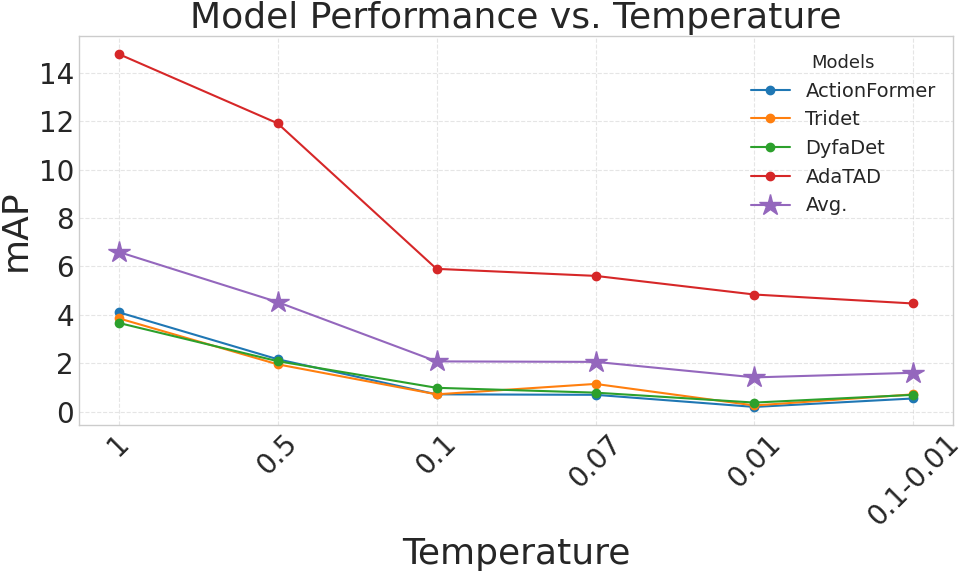}
    \caption{The influence of \textit{temperature} in bidirectional contrastive loss. }
    \label{ana_trans}
\end{figure}
\subsection{The influence of temperature in Bi-con} 
As shown in Fig.~\ref{ana_trans}, when the temperature value undergoes a decrement from 1 to 0.01. The last column represents the temperature $\tau$ gradually decays from 0.1 to 0.01 in the iterative training of the perturbation. The decay follows an exponential schedule. There is a notable and distinct drop in the performance of the models. This performance decline is concomitant with an observed enhancement in attack strength. 

This behavior stems from the fact that temperature controls the sharpness of the similarity distribution: smaller values produce a more peaked distribution, thereby improving feature discrimination. In turn, this effect places stronger emphasis on hard negatives, enabling the generation of more transferable adversarial examples. It is worth noting that the most favorable results occur when the temperature is set between 0.01 and 0.07, suggesting that moderately low values provide an effective balance between attack robustness and training stability.

\subsection{Resistance to augmentations}
\begin{table}[]
\begin{tabular}{l|ccccc}
\toprule
                  & AF & Tridet & DyfaDet & AdaTAD & Avg. \\
                  \midrule
MI-FGSM            & 18.2         & 14.2   & 16.6    & 31.2   & 20.1 \\
\textbf{Ours}              & 10.6         & 9.3    & 7.6     & 24.7   & 13.1 \\
\midrule
MI-FGSM & 11.9         & 10.4   & 15.9    & 23.2   & 15.4 \\
\textbf{Ours}   & 7.7          & 6.8    & 6.9     & 18.9   & 10.1 \\
\bottomrule
\end{tabular}
\caption{Results of our attack and baselines under both standard and adaptive Expectation-over-Transformation (EOT) defenses on the THUMOS14 dataset. AF stands for ActionFormer.}
\label{tab:d}
\end{table}
Table~\ref{tab:d} presents the results under augmentation-based defenses. The first two rows correspond to non-adaptive augmentation defenses (temporal jittering, frame dropping, smoothing, and resampling), while the last two rows represent the adaptive EOT variant, where the defense is aware of the applied transformations during attack optimization. Our method achieves consistently lower average mAP across all detectors, indicating stronger attack effectiveness. Even under adaptive EOT defenses, our attack remains superior, showing enhanced robustness and transferability against transformation-aware defenses.

\subsection{Analysis of attack efficiency}
\begin{table}[ht]
    \centering
    \begin{tabular}{c|>{\centering\arraybackslash}p{3cm}|c}
    \toprule
    Method & Num. of samples needed per iteration & Avg. time (s) \\
    \midrule
     I-FGSM    & 1 & 10.17 \\
      MI-FGSM   &  1 & 9.83\\
      DI-FGSM & 1 & 10.33\\
      TI-FGSM & 1 & 9.83 \\
      SIM & 5 & 44.00\\
      BSR & 3 & 26.83 \\
      \midrule
      Our TVA & 1 & 9.83\\
      \bottomrule
    \end{tabular}
    \caption{Analysis of the attack efficiency.}
    \label{tab:eff}
\end{table}
We evaluate the real-time efficiency of the aforementioned attack methods by measuring the average time taken to generate a single adversarial example on SEEDBench, using the SigLIP model and an NVIDIA A40 GPU. The results are summarized in Table~\ref{tab:eff}.

Our method, TVA, demonstrates the best computational efficiency among all approaches. While augmentation-based attacks like BSR and SIM rely on multiple forward passes per iteration due to input transformations, TVA introduces a lightweight, plug-and-play loss that enables one-time gradient enhancement per iteration, thus preserving the best runtime cost.

\subsection{Implementation details} 
\subsubsection{Experimental setting.}
For TAD, we follow the attack settings described as Form(b) in Page 4 of the Main paper, namely, frozen backbone, when evaluating ActionFormer, TriDet, and DyfaDet. In contrast, AdaTAD is evaluated under Form(a), where the backbone is fine-tuned together with the task head. The temperature parameter is fixed at $\tau=0.01$ for all TAD experiments.
For all other downstream tasks, we adopt Form(b) (frozen backbone) consistently across all models for practical deployment efficiency. In this setting, we use a temperature annealing schedule, decaying $\tau$ from $0.1$ to $0.005$ over the course of training perturbations.
Each model is evaluated using its original training hyperparameters and data preprocessing pipeline, ensuring comparability with existing results.

\subsubsection{Hardware Setup.}
TAD experiments are conducted using a server equipped with 4 NVIDIA A40 GPUs, whereas all other attack experiments are performed on a single NVIDIA A40 GPU.

\end{document}